\documentclass[9pt, conference, a4paper, compsocconf]{IEEEtran}
\IEEEoverridecommandlockouts
\usepackage{amsmath,amssymb,amsthm,amsfonts}
\usepackage[top=2.7cm, bottom = 5.4cm, left=2.72cm, right = 2.85cm]{geometry}
\usepackage{fancyhdr}
\usepackage{cite}
\usepackage{graphicx}
\usepackage{tabularx}
\usepackage[table]{xcolor}
\usepackage{multirow}
\usepackage[english]{babel}
\usepackage{array}

\usepackage[T5]{fontenc}
\usepackage[utf8]{inputenc}
\usepackage[utf8]{vietnam}

\makeatletter
\AtBeginDocument{%
  \renewcommand{\refname}{References}%
}
\makeatother

\makeatletter
\AtBeginDocument{
  \renewcommand{\abstractname}{Tóm tắt}
}
\renewcommand{\IEEEkeywordsname}{Từ khóa}
\makeatother

\usepackage{caption}
\usepackage[font=footnotesize,caption=false]{subfig}

\makeatletter
\def\@maketitle{%
  \newpage
  \null
  \begin{center}%
    {%
      \fontsize{16}{18}\selectfont
      \bfseries \@title \par
    }%
    \vskip 4em%
    {%
    \fontsize{9}{3.5}\selectfont
      
      \begin{tabular}[t]{c}%
        \@author
      \end{tabular}\par
    }%
  \end{center}%
  \par
}

\renewenvironment{abstract}
  {\normalfont
   \@IEEEabskeysecsize
   \setlength{\parindent}{0pt}
   \noindent
   \if@twocolumn
     \bfseries\textit{\abstractname}: 
   \else
     \begin{center}\textbf{\abstractname}\end{center}
   \fi}
  {\par}
  
\renewenvironment{IEEEkeywords}
  {\normalfont
   \@IEEEabskeysecsize
   \vspace{0.3em}       
   \setlength{\parindent}{0pt}
   \noindent
   {\bfseries\itshape \IEEEkeywordsname: }\ignorespaces}
  {\par}

\def\IEEEbibitemsep{1.5pt plus .5pt}

\makeatother

\newif\ifrevision
\revisionfalse     

\ifrevision
  \newcommand{\rev}[1]{\colorbox{yellow!25}{\strut #1}}
\else
  \newcommand{\rev}[1]{#1}
\fi

\ifrevision
  \newcommand{\revblock}[1]{%
    \begingroup
    \setlength{\fboxsep}{4pt}%
    \colorbox{yellow!15}{\parbox{\linewidth}{#1}}%
    \endgroup
  }
\else
  \newcommand{\revblock}[1]{#1}
\fi

\ifrevision
  \newcommand{\revtablebg}{\rowcolors{0}{yellow!12}{yellow!12}}
\else
  \newcommand{\revtablebg}{}
\fi

\usepackage{microtype}

\usepackage[hidelinks]{hyperref}
\hypersetup{pageanchor=false}

\begin{document}

\pagenumbering{gobble}

\title{AI Agents for Sustainable SMEs:\\ A Green ESG Assessment Framework}

\author{
\IEEEauthorblockN{
    Viet Trinh\IEEEauthorrefmark{1}\IEEEauthorrefmark{2}, 
    Tan Nguyen\IEEEauthorrefmark{1}\IEEEauthorrefmark{2}, 
    Minh-Huyen Phan\IEEEauthorrefmark{1}\IEEEauthorrefmark{2}, 
    and Quan Luu\IEEEauthorrefmark{1}\IEEEauthorrefmark{2}
}
\IEEEauthorblockA{\IEEEauthorrefmark{1}Faculty of Information Systems, University of Economics and Law}
\IEEEauthorblockA{\IEEEauthorrefmark{2}Vietnam National University, Ho Chi Minh, Vietnam}
\IEEEauthorblockA{Email: tqviet@uel.edu.vn, tanndm22416c@st.uel.edu.vn,\\ huyenptm22416c@st.uel.edu.vn, quanlm22416c@st.uel.edu.vn}
}

\maketitle

\thispagestyle{IEEEtitlepagestyle}

\begin{abstract}
This study presents a novel, AI-driven framework for assessing Environmental, Social, and Governance (ESG) performance in European small and medium-sized enterprises (SMEs). An initial phase established expert-validated ESG baseline scores from a subset of the Flash Eurobarometer FL549 survey data. In the second phase, a scalable AI agent system, built on the n8n automation platform, applied these baselines to perform automated ESG classification and generate contextual recommendations using large language models (LLMs). The results demonstrate the AI system's high consistency with human-derived outputs, thereby supporting more effective monitoring and intervention strategies aligned with the European Green Deal.
\end{abstract}


\section{Introduction} \label{sec-introduction}
The green and digital transitions, collectively referred to as the \textit{twin transition}, have become central to the European Union’s (EU) policy and economic agenda, particularly in the context of small and medium-sized enterprises (SMEs). As the backbone of the European economy, SMEs account for over 99\% of all businesses and play a pivotal role in driving sustainable growth, innovation, and job creation \cite{European2025Report}. However, their relatively limited access to technical resources and policy guidance poses challenges in aligning with ambitious Environmental, Social, and Governance (ESG) standards. In this context, developing scalable and intelligent frameworks to evaluate and enhance SMEs’ green performance has emerged as a crucial priority for policymakers and researchers alike.

Previous literature has investigated the environmental engagement of SMEs from various disciplinary perspectives, identifying barriers, motivations, and patterns of eco-innovation. For example, Rizos et al. \cite{Rizos2016Implementation} highlighted the limitations of internal capacity and the lack of access to green finance as the main obstacles that prevent SMEs from adopting circular business models. The study by Garrido et al. \cite{Garrido2024Implementation} provides partial evidence that the implementation of ESG in small businesses, particularly labor practices, can improve financial performance and enhance resilience during crises. More recently, Chen et al. \cite{Chen2024Influence} indicates that digital technologies, including artificial intelligence (AI), machine learning, and blockchain, can enhance the quality and assurance level of ESG reporting by automating data collection, improving consistency, and minimizing green-washing. Despite these promising developments, existing solutions often remain top-down, insufficiently tailored to SME realities, or lacking real-time recommendation capabilities based on structured survey data.

To address these limitations, this paper introduces a novel assessment framework that leverages AI agents to evaluate the ESG performance of European SMEs. Whereas FL549 survey collects the answers at the single level of the SMEs, the data released to the population is at the national scale. Using data from the \textit{Flash Eurobarometer FL549} survey, this study employs a two-stage methodology, with every state being considered as a representative of a compounded SME ecosystem. The first stage involves a data processing pipeline that extracts and aggregates ESG-related indicators from structured survey data across four key groups: \textit{resource efficiency}, \textit{circular energy}, \textit{climate strategy}, and \textit{green product offerings}. The second stage utilizes an AI agent system, built on the \textit{n8n} automation platform, to perform automated classification and generate contextual recommendations based on country-level ESG scores. This framework thus provides a transparent benchmarking tool and a decision assistant that leverages large language models for adaptive recommendations.

\section{Related Works} \label{sec-related}
A review of recent literature highlights the critical role of European SMEs in the green transition, particularly concerning their ESG performance and resource efficiency. Chatzistamoulou et al. \cite{Chatzistamoulou2022} analyzed data from over 37,000 EU SMEs and found that a commitment to resource efficiency is a key driver of sustainability trends. Their findings suggest that factors such as technology adoption, collaboration, and expert advice, alongside resource productivity and green energy, significantly enhance this commitment. Similarly, Sanz-Torró et al. \cite{SanzTorro2025} utilized a DEA-Bootstrap approach on a dataset of 13,343 European SMEs to demonstrate an asymmetry between green marketing efforts and actual resource efficiency improvements. The study indicates that while green marketing has advanced, resource efficiency has lagged, emphasizing the need for additional support, such as assessment tools, consulting, and funding. Furthermore, De Andrade et al. \cite{DEANDRADE2025144477} grouped 61,086,268 SMEs across 36 European countries based on their sustainability practices using the Flash Eurobarometer database. The study identified four distinct groups, ranging from firms with minimal practices (e.g., waste minimization) to those with advanced strategies (e.g., prioritizing water, energy, and material savings). The results underscore the importance of both investing in resource efficiency and fostering "green jobs" to promote sustainable practices within the SME sector.

Further research on ESG implementation and green economy challenges for SMEs highlights a range of difficulties. Numerous studies identify that SMEs encounter political, technical, economic, and market barriers when implementing sustainable practices. For instance, De Andrade et al. \cite{DEANDRADE2025144477} note that, even with supportive policies, SMEs face significant internal limitations, including a lack of human resources, professional knowledge, and sufficient financing to fully execute environmental actions. Similarly, a study by Bayisbayev et al. \cite{Bayisbayev2025} on ESG mapping in SMEs demonstrates that factors such as leadership commitment, regulatory compliance, stakeholder engagement, and financial stability are decisive in a firm's ability to implement ESG initiatives. The authors caution that a fragmented approach, such as solely pursuing a short-term "carbon neutral" goal, may not yield immediate financial benefits. From a policy perspective, Jin et al. \cite{Jin2025} recommend that post-COVID-19 governments in OECD countries should implement comprehensive policies to promote ESG investment and support SMEs in transitioning to clean energy.

The academic discourse on ESG assessment frameworks frequently identifies a lack of common standards. For instance, da Cunha et al. \cite{daCunha2025} systematically synthesized key ESG indicators and proposed a conceptual framework linking these components to corporate performance. They argued that the lack of regulatory standards for ESG disclosure leads to heterogeneous reporting structures, which diminishes comparability and necessitates the establishment of standardized benchmarks. Similarly, the OECD \cite{oecd2025esg} emphasizes that ESG measures must be "complete, comparable and meaningful", but notes that current ESG data is often inconsistent due to varying definitions and measurement scopes across standards and providers. This inconsistency creates "cumulative inconsistencies" in ESG scores, complicating efforts to prioritize critical performance dimensions. In response to these challenges, a number of studies have proposed models for ranking ESG strategies using multi-criteria decision-making methods, such as fuzzy TOPSIS integrated with Artificial Intelligence, as demonstrated by Aljohani et al. \cite{Aljohani2025}. These approaches enable the calculation of objective criterion weights and help to mitigate ambiguity in the evaluation process.

Emerging studies have increasingly focused on the application of AI and agent-based models for automated ESG assessment. For instance, Lee et al. \cite{Lee2024} developed the ESG2PreEM framework, which utilizes machine learning with Natural Language Processing (NLP) to automatically score ESG performance from news texts. The framework, which combines BERT, RoBERTa, and ALBERT models trained on LexisNexis data, achieved an impressive 80.8\% accuracy in classifying ESG documents and demonstrated consistency with MSCI scores. Similarly, Katsamakas et al. \cite{Katsamakas2023} employed an agent-based computational model to simulate market competition when firms invest in ESG. Their findings indicate that if ESG investment only attracts consumers, it has a negligible effect on competition. However, when firms leverage ESG to innovate products or processes, it intensifies competition, leading to lower prices and reduced profits. Additionally, research by Gao et al. \cite{Gao2024} explored the integration of large language models (LLMs) into agent simulations. They concluded that LLMs enable virtual agents to make more complex, human-like decisions and demonstrate greater adaptability to their environment, thus enhancing their utility in sophisticated modeling tasks.

\section{Methodology} \label{sec-method}
\begin{figure*}[!t]
\centering
\includegraphics[width=0.75\linewidth]{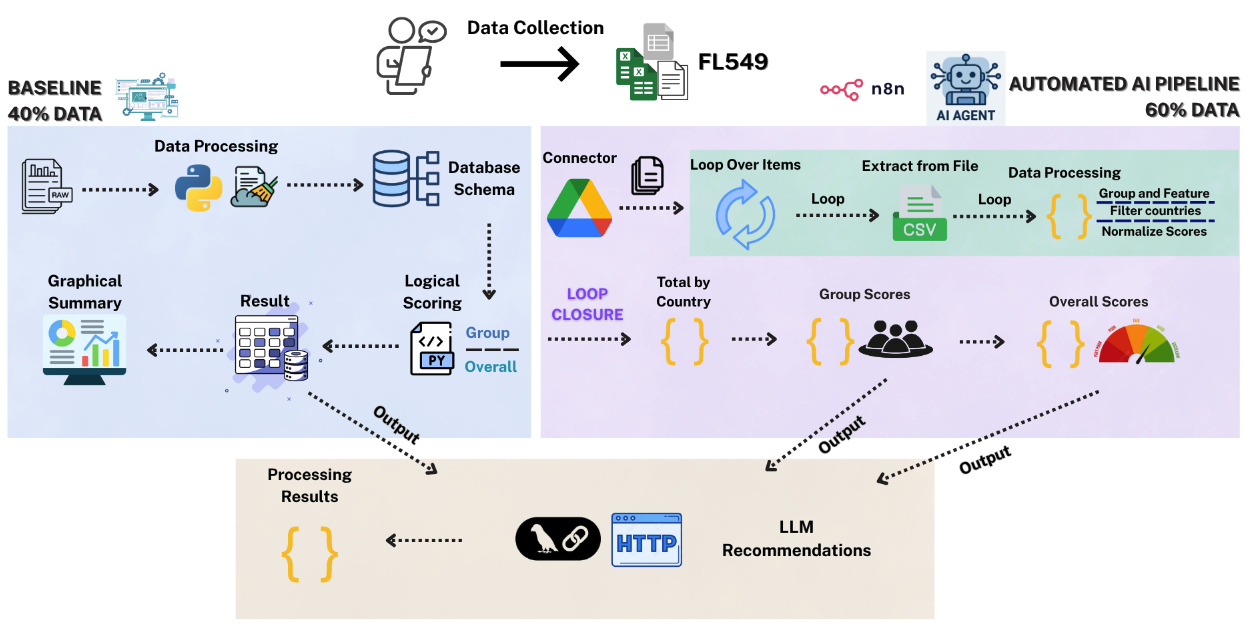}
\caption{A Human-AI workflow for establishing an ESG baselines and performing automated ESG analytics.}
\label{fig:pipeline}
\end{figure*}

This study proposes a hybrid human-AI orchestration framework, which begins with a human-curated baseline and culminates in a generative AI-augmented analytical workflow, as illustrated in the Figure~\ref{fig:pipeline}. The initial phase involves manually processing approximately 40\% of the \textit{Flash Eurobarometer FL549} survey data to establish a normative reference for subsequent automated inference. The remaining dataset is then processed through an automated pipeline orchestrated by the \textit{n8n} platform. This pipeline uses domain-driven heuristics to classify each data point into a composite \textit{ESG score}. An LLM agent, via prompt engineering, then generates interpretive narratives on score distributions, benchmarking context, and trend signals. This methodology provides a scalable, traceable, and scientifically grounded architecture for ESG benchmarking in the SME sector, effectively blending statistical rigor with adaptive AI-infused interpretation.

This study adopt a 40/60 baseline/automated split to balance \emph{tier-boundary stability} and \emph{evaluation coverage} in a small-$n$ country setting. A smaller baseline fraction makes quartile cut-points volatile, while a larger baseline reduces the automated evaluation set.
To mitigate concerns that results depend on a single random split, the study specifies a \emph{Repeated Random Sub-sampling Validation (RRSSV)} protocol over $S$ random seeds: for each seed, it resamples the 40\% baseline set, recompute tier thresholds, execute the automated workflow on the remaining countries, and compute evaluation metrics. Stability by mean and standard deviation across seeds are calculated as:

\begin{equation}
\bar{m}=\frac{1}{S}\sum_{s=1}^{S} m_s,\qquad
s_m=\sqrt{\frac{1}{S-1}\sum_{s=1}^{S}(m_s-\bar{m})^2}
\end{equation}

\subsection{Data Collection and Pre-processing}
The primary dataset for this study is the \textit{Flash Eurobarometer 549: SMEs, Resource Efficiency and Green Markets}, released by the European Commission in 2024. This survey offers a comprehensive overview of environmental sustainability practices among SMEs across the European Union, focusing on areas such as resource efficiency, circular economy, renewable energy, and green product development. Specifically, this research utilizes \textit{Volume A: Countries/EU}, which provides aggregated indicators at both country and EU levels, including weighted response frequencies and basic bivariate statistics. The dataset is highly representative, encompassing over 28,000 SMEs from 27 EU Member States and associated countries. The survey employs a standardized questionnaire with both categorical and ordinal variables, ensuring cross-national comparability. To maintain proportional representation, each enterprise response was weighted based on sector, firm size, and region. 

Besides the 27 European Union Member States, the Flash Eurobarometer FL549 survey also covers some non-EU and affiliated economies, including the United Kingdom, the United States as well as some of the European partner countries. Although the survey was initially conducted on an SME level, the one to be used in this study provides country-level aggregated indicators. This then implies that all further analysis will be done at the national scale whereby each individual country ESG profile is an integrated collection of SME sustainability practices in the respective jurisdiction which will in turn allow comparative analysis of the entire geographical scope of the FL549 dataset.

\subsubsection{Initial Treatment}
The data pre-processing phase focused on transforming the raw survey data into a structured and analysis-ready format. The dataset, originally in an Excel workbook, contained multiple sheets corresponding to distinct survey questions. To prepare the data, the first ten rows of each sheet were systematically skipped to remove non-tabular metadata. Subsequently, all completely empty rows and columns were eliminated to reduce noise and enhance data integrity. Rows containing textual question identifiers, typically enclosed in parentheses, were also filtered out as they did not represent quantitative responses. A crucial normalization step involved forward-filling empty cells in the criteria column using the most recent valid entry, thereby preserving semantic consistency for downstream processing. 

\subsubsection{Data Aggregation and Categorization} \label{sec:data-category}
Responses to the survey were categorized into four principal types, each requiring a distinct processing strategy to generate meaningful and comparable country-level scores.

\begin{itemize}
    \item \textbf{\textit{Single-Choice Responses}}: for questions allowing only one answer, each option was assigned a score on a 0 - 10 ordinal scale, reflecting the option's semantic polarity, with 0 representing the most negative and 10 representing the most advanced one.
    \item \textbf{\textit{Multiple-Choice Responses}}: for questions allowing multiple answers, each selected option was scored individually. The country-level score was then calculated as a weighted average of the individual option scores, aggregated across all respondents in that country.
    \item \textbf{\textit{Maximum Multiple-Choice Responses}}: for questions allowing multiple answers with a specified maximum, the country-level score is also the weighted average of the individual option scores, but with special constraint on the maximum number of responses allowed per participant.
    \revblock{
    \item \textbf{\textit{WRITE-DOWN (binned numeric) Responses}}: although labeled as ``write-down'' in the original questionnaire, the only such variable used in our scoring (DX5: employees in green jobs) is fully discretized into predefined bins during data collection and cleaning (e.g., 0 employees, 1--5, 6--9, 10--50, 51--100, 101+, and dk/na). Therefore, no free-text processing or lexicon is used. The country-level score is computed from the binned response distribution and then normalized to a 0--10 scale.
    }
\end{itemize}

\subsection{ESG Groups and Baseline Scores}
To facilitate a comparative ESG analysis across countries, scores were computed for four thematic groups: \textit{Governance}, \textit{Energy \& Circular Economy}, \textit{Biodiversity}, and \textit{Climate Strategy}. This procedure began by applying group-specific importance weights and then standardizing the score distributions to ensure cross-group comparability. A linear min-max scaling was applied independently within each group, mapping the weighted scores onto a fixed range of 1 to 10. For each group, this study evaluated key descriptive statistics (mean, standard deviation, and quartiles) and conducted formal statistical tests to assess the distribution of scores. The Shapiro-Wilk test was used for smaller samples, while the D’Agostino-Pearson omnibus test was applied to detect deviations from normality in larger samples. A group is considered to approximate a normal distribution only if both tests yielded a p-value greater than 0.05.

\subsection{Scoring Formalization and Reproducibility} \label{sec:scoring-formalization}
\textbf{Notation.} Let $c \in \mathcal{C}$ denote a country, $q \in \mathcal{Q}$ a survey indicator (question), and $o \in \mathcal{O}_q$ a response option/bin. Let $f_{c,q,o}$ be the (weighted) response frequency (or percentage) of option $o$ for country $c$. Each option is assigned an expert-defined score $s_{q,o} \in [0,10]$ reflecting semantic polarity (0: least advanced; 10: most advanced).

\textbf{Indicator score.} For each indicator $q$, the country-level score is computed as the expected score under the observed response distribution:
\begin{equation}
x_{c,q}=\frac{\sum_{o\in\mathcal{O}_q} f_{c,q,o}\, s_{q,o}}{\sum_{o\in\mathcal{O}_q} f_{c,q,o}}.
\end{equation}

\textbf{Pillar score.} For each ESG pillar $g \in \{\mathrm{GOV},\mathrm{ENE},\mathrm{BIO},\mathrm{CLI}\}$ with indicator set $\mathcal{Q}_g$, we compute the pillar score as an unweighted mean over indicators:
\begin{equation}
S_{c,g}=\frac{1}{|\mathcal{Q}_g|}\sum_{q\in\mathcal{Q}_g} x_{c,q}.
\end{equation}

\textbf{Weighted ESG aggregation.} A composite national ESG score is computed by fixed normative pillar weights:
\begin{equation}
\begin{aligned}
\mathrm{ESG}_c &= \sum_{g} w_g\, S_{c,g},\\
(w_{\mathrm{GOV}}, w_{\mathrm{ENE}}, w_{\mathrm{BIO}}, w_{\mathrm{CLI}})
&= (0.1,\,0.5,\,0.3,\,0.1).
\end{aligned}
\end{equation}

\textbf{WRITE-DOWN (DX5) bin normalization.} DX5 is treated as a binned numeric indicator. Let midpoints be $m_o \in \{0,3,7.5,30,75,120\}$ for bins $\{0,1$--$5,6$--$9,10$--$50,51$--$100,101+\}$, and define
\begin{equation}
s_{\mathrm{DX5},o}=10\cdot \frac{m_o-\min(m)}{\max(m)-\min(m)}.
\end{equation}

\subsection{AI-Agent Workflow}
\begin{figure*}[!t]
\centering
\includegraphics[width=0.8\linewidth]{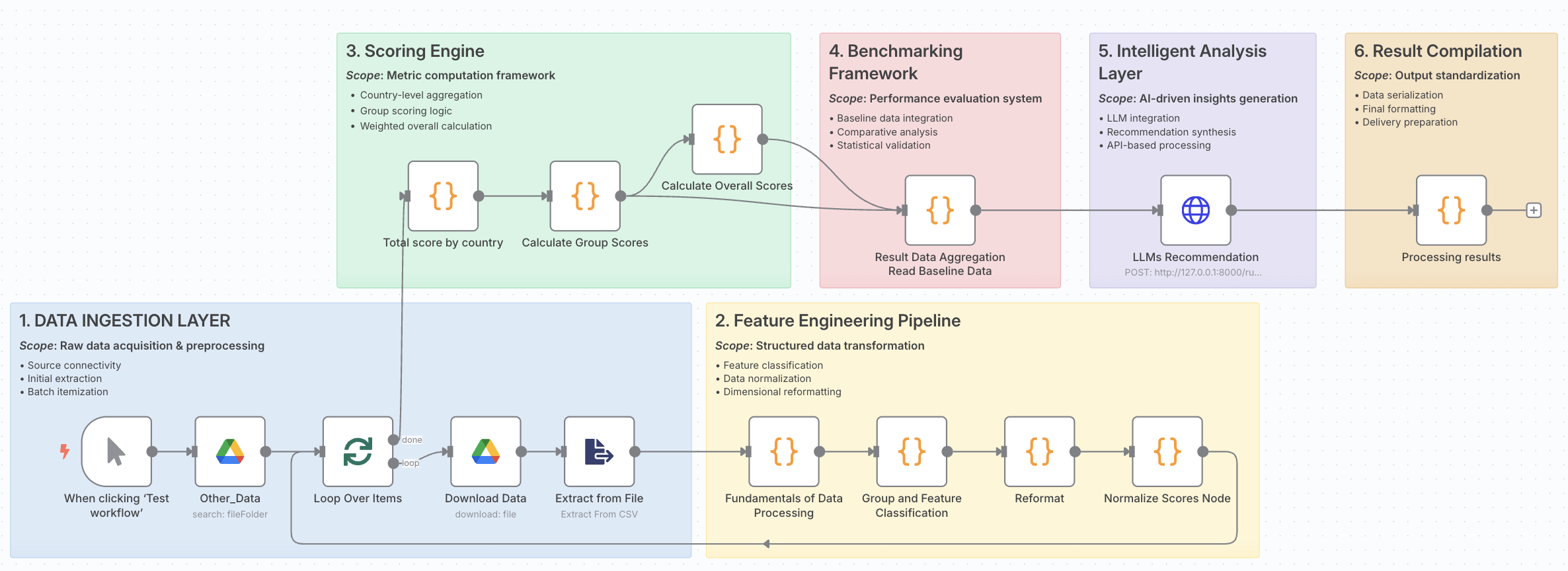}
\caption{An end-to-end AI-agent workflow for ESG analytics.}
\label{fig:workflow}
\end{figure*}

The AI-Agent workflow is structured to process and benchmark new data against established ESG baselines (Figure~\ref{fig:workflow}). Data ingestion is handled by a subsystem that retrieves information in an iterative, batch-wise manner to maintain memory efficiency and enable modular file processing. An initial extraction phase infers the data schema and validates essential metadata. The subsequent data transformation is performed by a sequence of four custom processing nodes:
\begin{enumerate}
    \item \textbf{\textit{Standardization}}: a processing node standardizes file codes for consistency.
    \item \textbf{\textit{Classification}}: a dedicated node assigns each question to one of four principal types based on pre-validated domain rules, as described in Section~\ref{sec:data-category}.
    \item \textbf{\textit{Filtering}}: a filtering node retains only the data associated with the designated countries for analysis.
    \item \textbf{\textit{Normalization}}: a normalization node computes scaled scores (on a 0-10 scale) by converting raw counts and ratios based on the logic of each feature type.
\end{enumerate}

Following this transformation, data is aggregated through a three-stage score computation process: combining normalized responses, computing average weights, and aggregating these into a composite national score. The final step involves a communication node that interfaces with an LLM service to generate interpretive narratives of the quantitative results using prompt engineering techniques.

\section{Results and Discussion} \label{sec-results}
For consistency, the study uses the terms \emph{pillar} and \emph{group} interchangeably to refer to the four ESG dimensions. The ESG ratings, classifications, and comparative analysis data engaged in this work are performed at the national level. Although the underlying data is based on SME-level survey data, the indices that are created are country-level ESG indices, thus, capturing aggregate sustainability trends dominating national SME ecosystems, and not the performance of individual firms.

\subsection{Statistical Analysis of ESG Baseline Scores}
\begin{itemize}
    \item \textit{Group 1 - Governance}: with 120 observations, the mean score for this group is 4.48 (standard deviation = 2.17) on a 10-point scale, indicating a moderate-to-low level of performance with significant score variability. The median of 3.87, which is lower than the mean, suggests a right-skewed distribution, driven by a small number of high scores. The interquartile range of 2.51 to 6.31 shows that the middle 50\% of scores are concentrated within this range. As both the Shapiro-Wilk and D'Agostino-Pearson tests yielded p-values $<$ 0.05, the data for this group do not follow a normal distribution.
    \item \textit{Group 2 - Energy \& Circular Economy}: with 60 observations, the mean score for this group is 3.81 (standard deviation = 2.59) on a 10-point scale, indicating a relatively low central tendency with significant score dispersion. The median of 2.50, which is lower than the mean, suggests a right-skewed distribution influenced by a few high-scoring outliers. The interquartile range spans from 1.70 to 6.23, highlighting a wide spread for the middle 50\% of the data. As both the Shapiro-Wilk and D'Agostino-Pearson tests yielded p-values $<$ 0.05, the data for this group also do not follow a normal distribution.
    \item \textit{Group 3 - Biodiversity}: with 30 observations, the mean score for this group is 4.48 (standard deviation = 3.44), indicating a low central tendency but with significant score dispersion. The median of 2.50 is considerably lower than the mean, while the third quartile approaches 8, suggesting a strong right-skewed distribution influenced by high-scoring outliers. The large discrepancy between the mean and median, coupled with a high standard deviation, confirms a high degree of variability in assessments. As both tests yielded p-values $<$ 0.05, the study rejected the null hypothesis that the data follows a normal distribution.
    \item \textit{Group 4 - Climate Strategy}: with 105 observations, the mean score for this group is 3.42 (standard deviation = 2.29), indicating a low central tendency with a moderate degree of score dispersion. The median of 2.14 is lower than the mean, suggesting a right-skewed distribution influenced by a few high-scoring outliers. The interquartile range spans from 1.56 to 5.16, indicating that the central 50\% of scores are concentrated in the low-to-medium range. As both tests yielded p-values $<$ 0.05, the study also rejected the null hypothesis that the data follows a normal distribution.
\end{itemize}

\subsection{Systematic Agreement Evaluation} \label{sec:agreement-eval}
In this study, the workflow outputs were evaluated against expert baselines using standard agreement metrics at both continuous and categorical levels. For each pillar $g$, mean absolute error (MAE), root mean squared error (RMSE), signed bias, and Spearman rank correlation between baseline scores $B_{c,g}$ and workflow scores $W_{c,g}$ are defined. Quartile labels (Weak/Average/Good/Excellent) are derived from the baseline distribution and applied consistently to both baseline and workflow for categorical agreement metrics (Accuracy, Macro-F1, and Cohen's $\kappa$).

\begin{equation}
\begin{aligned}
\mathrm{MAE}_g
&=\frac{1}{|\mathcal{C}|}\sum_{c\in\mathcal{C}} \left| W_{c,g}-B_{c,g}\right|,\\
\mathrm{RMSE}_g
&=\sqrt{\frac{1}{|\mathcal{C}|}\sum_{c\in\mathcal{C}} \left(W_{c,g}-B_{c,g}\right)^2 }.
\end{aligned}
\end{equation}

\subsection{Learning-based Baseline (Design Specification)} \label{sec:ml-baseline}
\begin{figure}[!t]
    \centering
    \subfloat[a]{%
        \includegraphics[width=0.44\linewidth]{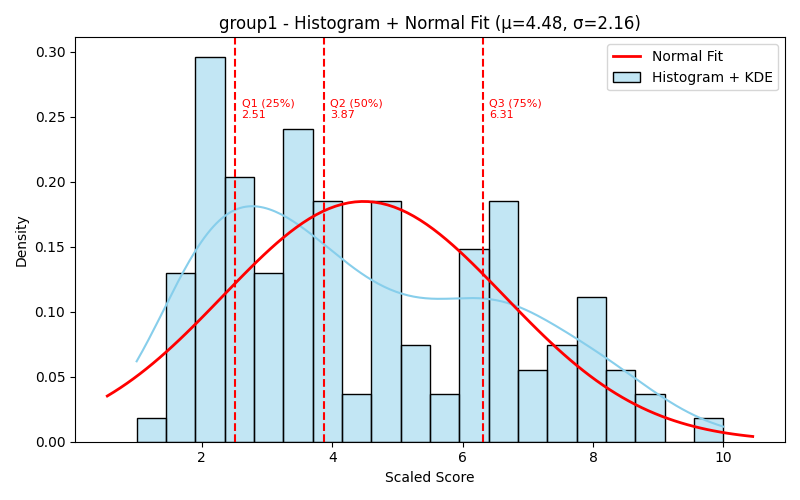}%
        \label{fig:baseline-g1}
    }
    \hfill
    \subfloat[b]{%
        \includegraphics[width=0.44\linewidth]{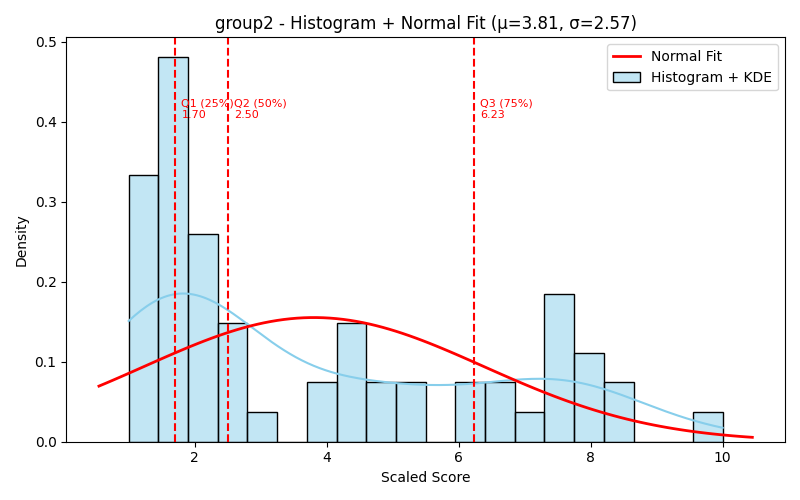}%
        \label{fig:baseline-g2}
    }
    \vspace{0.18cm}
    \subfloat[c]{%
        \includegraphics[width=0.44\linewidth]{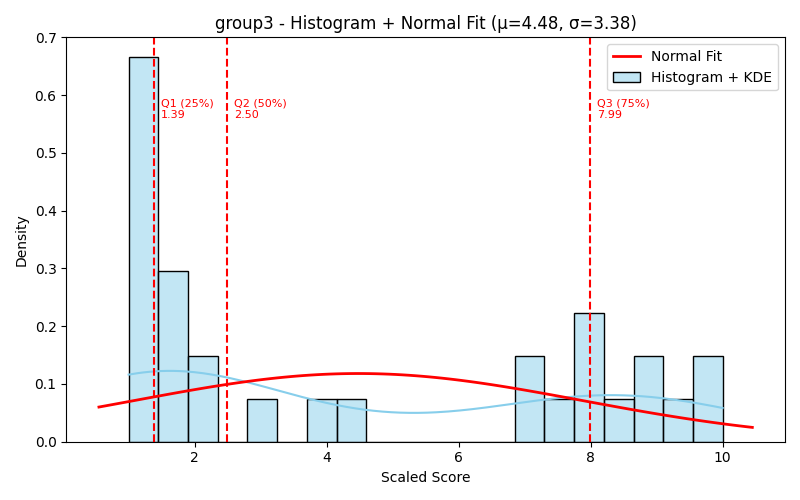}%
        \label{fig:baseline-g3}
    }
    \hfill
    \subfloat[d]{%
        \includegraphics[width=0.44\linewidth]{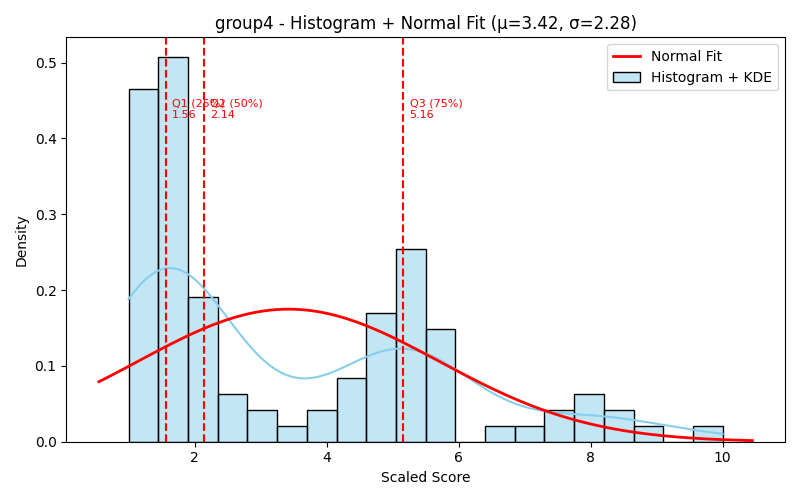}%
        \label{fig:baseline-g4}
    }
    \caption{Baseline score distributions for four groups:
    (a) Group 1 - Governance,
    (b) Group 2 - Energy and Circular Economy,
    (c) Group 3 - Biodiversity,
    (d) Group 4 - Climate Strategy. (Source: Authors)} 
    \label{fig:baseline-distribution}
\end{figure}

Reviewers suggested adding a learning-based comparator. Given the country-level aggregation and the absence of externally validated ESG ground-truth labels in FL549, the study specifies a minimal supervised baseline as a \emph{planned} experiment to contextualize the agentic pipeline without over-claiming predictive validity. Table~\ref{tab:ml_design} defines the baseline task, features, target, protocol, and metrics; quantitative results are reserved for future work when a robust label strategy and sufficient sample size are available. The baseline score distributions for each group are presented in Figure~\ref{fig:baseline-distribution}, and the results of the Shapiro-Wilk and D'Agostino-Pearson statistical tests for the four groups are shown in Table~\ref{tab:dist-test}.

\begin{table}[!t]
\centering
\caption{\rev{Planned learning-based baseline for ESG tier prediction (design specification).}}
\label{tab:ml_design}
\scriptsize
\setlength{\tabcolsep}{4pt}
\renewcommand{\arraystretch}{1.12}
\revtablebg
\begin{tabular}{|p{1.65cm}|p{4.95cm}|}
\hline
\textbf{Item} & \textbf{Specification} \\
\hline
Task & Predict tier label (Weak/Average/Good/Excellent) per country--group pair. \\
\hline
Features & Questionnaire-derived normalized signals (question-level scores prior to aggregation). \\
\hline
Target & Tier labels induced by baseline quartiles (group-specific). \\
\hline
Model & Multinomial logistic regression (L2 regularization) as minimal supervised baseline. \\
\hline
Train/Test & Train on baseline subset; test on held-out countries under the same split regime. \\
\hline
Validation & RRSSV with 40\% baseline resampling over $S$ seeds (protocol in Methodology). \\
\hline
Metrics & Accuracy and Macro-F1 (reported as mean $\pm$ std across seeds). \\
\hline
\end{tabular}
\rowcolors{1}{}{}
\end{table}

\begin{table}[!t]
  \centering
  \caption{The statistical distribution tests of four groups}
  \label{tab:dist-test}
  \scriptsize
  \setlength{\tabcolsep}{4pt}
  \renewcommand{\arraystretch}{1.2}

  \begin{tabular}{|p{2.7cm}|p{2.4cm}|c|}
    \hline
    \textbf{Group} & \textbf{Test} & \textbf{p-value} \\
    \hline
    \multirow{2}{*}{Governance}
      & Shapiro-Wilk          & 0.0000 \\
      & D’Agostino-Pearson    & 0.0005 \\
    \hline
    \multirow{2}{*}{Energy \& Circular Economy}
      & Shapiro-Wilk          & 0.0000 \\
      & D’Agostino-Pearson    & 0.0045 \\
    \hline
    \multirow{2}{*}{Biodiversity}
      & Shapiro-Wilk          & 0.0001 \\
      & D’Agostino-Pearson    & 0.0000 \\
    \hline
    \multirow{2}{*}{Climate Strategy}
      & Shapiro-Wilk          & 0.0000 \\
      & D’Agostino-Pearson    & 0.0033 \\
    \hline
  \end{tabular}
\end{table}

\subsection{Statistical Analysis of Classified ESG Scores}
\begin{figure}[!t]
    \centering
    \subfloat[a]{%
        \includegraphics[width=0.44\linewidth]{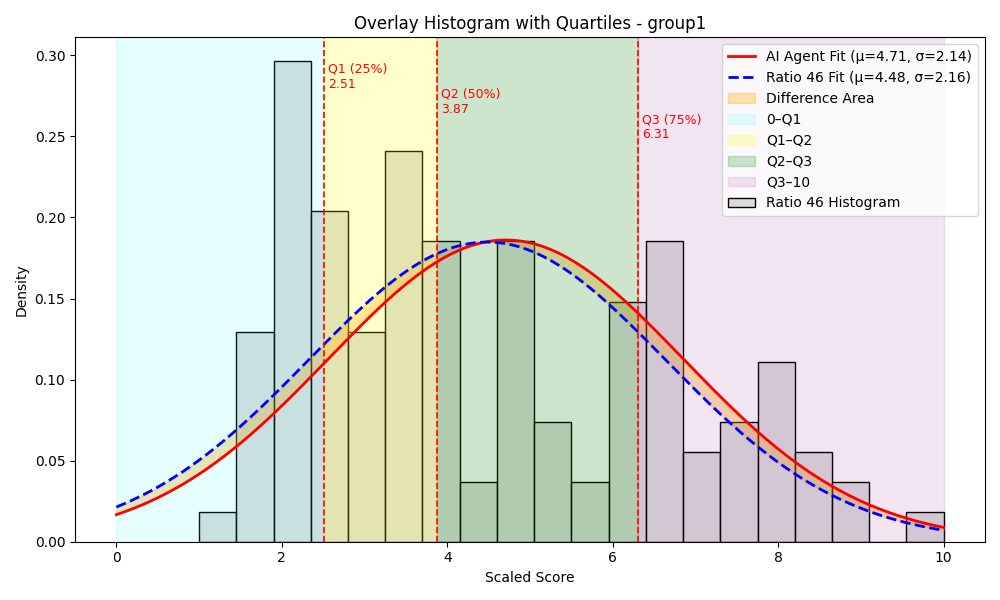}%
        \label{fig:overlay_g1}
    }
    \hfill
    \subfloat[b]{%
        \includegraphics[width=0.44\linewidth]{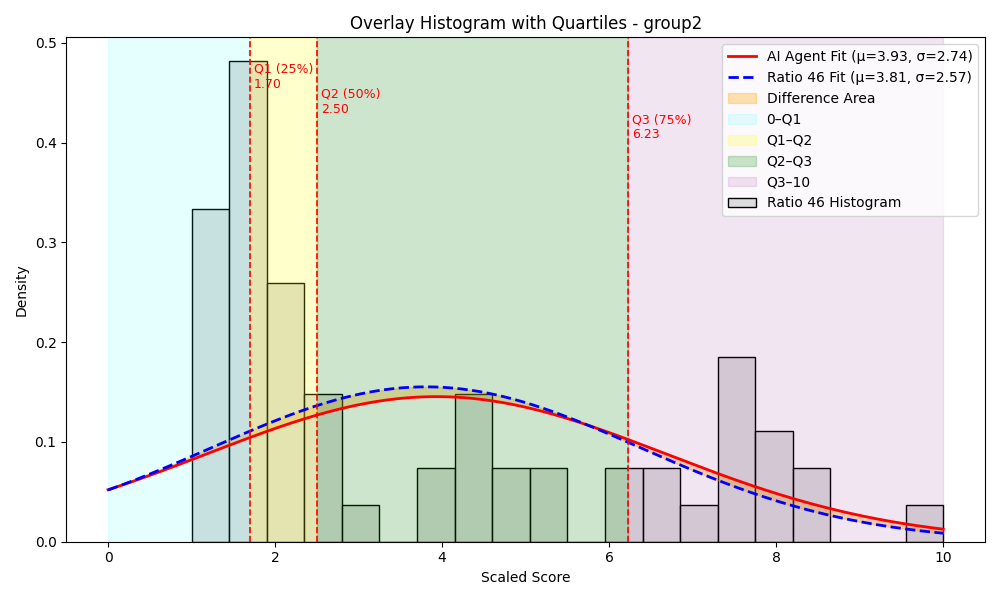}%
        \label{fig:overlay_g2}
    }
    \vspace{0.18cm}
    \subfloat[c]{%
        \includegraphics[width=0.44\linewidth]{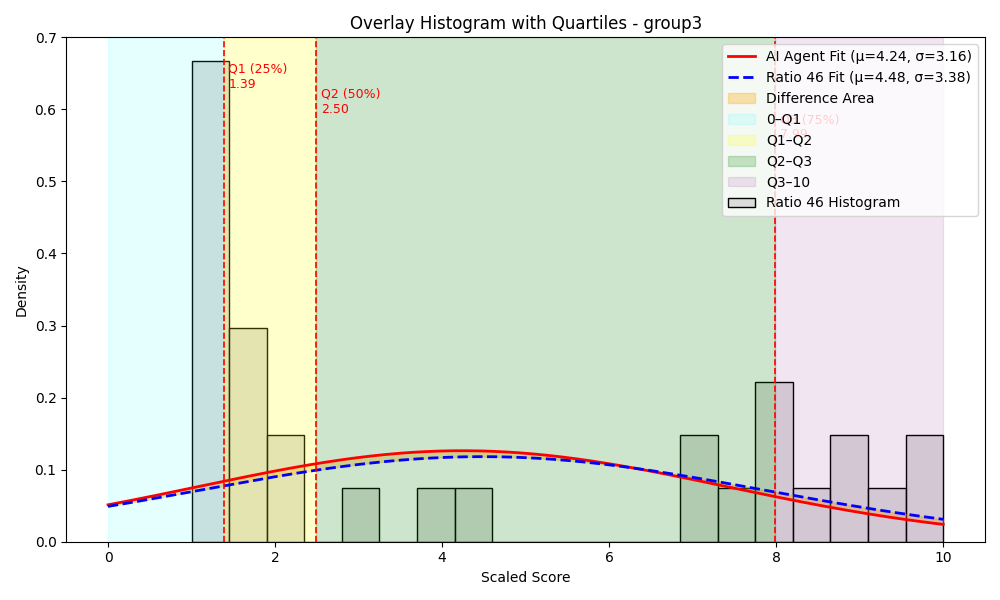}%
        \label{fig:overlay_g3}
    }
    \hfill
    \subfloat[d]{%
        \includegraphics[width=0.44\linewidth]{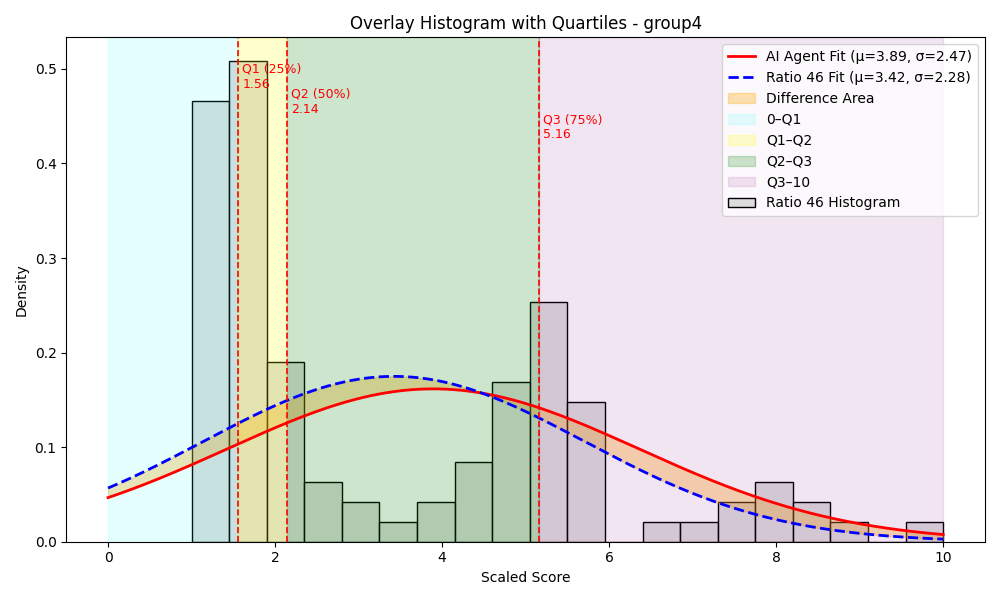}%
        \label{fig:overlay_g4}
    }
    \caption{Classified ESG score distributions for four groups:
    (a) Group 1 – Governance,
    (b) Group 2 – Energy and Circular Economy,
    (c) Group 3 – Biodiversity,
    (d) Group 4 - Climate Strategy. (Source: Authors)}
    \label{fig:overlay-distribution}
\end{figure}

Following classification by the AI-Agent workflow, each country's ESG score is benchmarked against the established baseline to assess its level of sustainability. This is achieved through a quartile-based classification system, as depicted in Figure~\ref{fig:overlay-distribution}, which segments countries into four performance categories. The first quartile (0 to Q1) is labeled as "\textit{Weak}", the second quartile (Q1 to Q2) as "\textit{Average}", the third quartile (Q2 to Q3) as "\textit{Good}", and the fourth quartile (Q3 to 10) as "\textit{Excellent}". This methodology identifies countries in the "Weak" and "Average" categories as candidates for targeted policy intervention to improve their ESG performance and rankings.

A direct comparison between the scores generated by the automated AI-Agent system and the expert-validated baseline scores is presented in the table~\ref{tab:overlay}. These numerical values highlight the differences, providing a clear and intuitive measure of consistency. In Group 1 - Governance, the scores generated by the AI-Agent system showed a high degree of concordance with the baseline across all four quartiles. Notably, the maximum deviation observed was approximately $-0.17$ in the "Good" region, which remained within the predefined tolerance and did not affect the overall classification. In contrast, Group 2 - Energy \& Circular Economy exhibited a notable positive bias. The AI-Agent workflow yielded significantly higher scores than the baseline in the Good ($+0.59$) and Excellent ($+0.5$) quartiles, suggesting a more optimistic evaluation in higher-scoring areas. Conversely, a minor negative deviation of approximately $-0.06$ was observed in the "Weak" region. Furthermore, Group 3 - Biodiversity demonstrated minimal and consistent deviations across all regions. The differences were negligible in the "Average" ($+0.01$) and "Excellent" ($+0.03$) quartiles. However, the largest deviation was a positive one in the "Good" region ($+0.65$), which indicates a tendency for the system to assign higher ratings in the mid-to-high performance range. Finally, in Group 4 - Climate Strategy, the system showed strong alignment with the baseline in the "Weak," "Average," and "Excellent" quartiles, with relatively small and consistent deviations ranging from $+0.03$ to $+0.26$. Nonetheless, a significant negative deviation of approximately $-0.49$ was observed in the "Good" region. Consequently, this discrepancy could potentially introduce a classification bias near the boundary between the "Good" and "Excellent" categories.

\begin{table*}[!t]
  \centering
  \caption{Score differences between AI-Agent workflow and baseline for all groups}
  \label{tab:overlay}
  \scriptsize
  \setlength{\tabcolsep}{4pt}
  \renewcommand{\arraystretch}{1.12}

  \begin{tabularx}{\textwidth}{|p{2.4cm}|p{2.2cm}|>{\centering\arraybackslash}X|>{\centering\arraybackslash}X|>{\centering\arraybackslash}X|}
    \hline
    \textbf{Group} & \textbf{Classification} & \textbf{Baseline} & \textbf{Workflow} & \textbf{Diff.} \\
    \hline
    \multirow{4}{*}{Governance}
      & Weak      & 2.033 & 1.997 & -0.036 \\
      & Average   & 3.279 & 3.194 & -0.085 \\
      & Good      & 5.112 & 4.945 & -0.167 \\
      & Excellent & 7.504 & 7.584 & 0.080 \\
    \hline
    \multirow{4}{*}{\shortstack{Energy \\ \& Circular}}
      & Weak      & 1.353 & 1.288 & -0.065 \\
      & Average   & 1.976 & 2.074 & 0.098 \\
      & Good      & 4.224 & 4.814 & 0.590 \\
      & Excellent & 7.676 & 8.211 & 0.535 \\
    \hline
    \multirow{4}{*}{Biodiversity}
      & Weak      & 1.198 & 1.251 & 0.053 \\
      & Average   & 1.697 & 1.708 & 0.011 \\
      & Good      & 5.908 & 6.557 & 0.649 \\
      & Excellent & 8.949 & 8.979 & 0.030 \\
    \hline
    \multirow{4}{*}{\shortstack{Climate \\ Strategy}}
      & Weak      & 1.290 & 1.326 & 0.036 \\
      & Average   & 1.760 & 1.813 & 0.053 \\
      & Good      & 3.964 & 3.474 & -0.490 \\
      & Excellent & 6.559 & 6.820 & 0.261 \\
    \hline
  \end{tabularx}
\end{table*}

\subsection{LLM-generated Contextual Recommendations}
Following the classification of nations into performance tiers, jurisdictions requiring enhanced sustainability measures are flagged for targeted policy intervention. The Gemini 2.0 Flash Lite large language model (LLM) is utilized to perform contextual reasoning and generate tailored ESG improvement recommendations. Representative examples of these generative outputs for underperforming nations are detailed in Table~\ref{tab:llm_suggestions}.

The country-level ESG performance is assessed by benchmarking scores against expert-validated baseline distributions. This methodology is prioritized to ensure transparency, interpretability, and alignment with policy-oriented benchmarking practices—domains where explainability and cross-country comparability take precedence over predictive optimization. While learning-based classifiers offer increased flexibility, their high data requirements and inherent complexity often obscure the interpretability required by policy stakeholders.

To evaluate the LLM-generated recommendations, an expert rubric is established based on three ordinal criteria (1–5): Relevance (alignment with diagnosed group-level gaps), Actionability (specificity and feasibility of prioritized steps), and Faithfulness (absence of unsupported claims or contextual contradictions). This framework ensures qualitative assessment reproducibility and explicitly penalizes hallucinatory behavior. Future iterations will involve multiple independent raters to report inter-rater reliability via Krippendorff's $\alpha$ alongside aggregated performance scores.

\begin{table}[!t]
\centering
\caption{\rev{Expert rubric for evaluating LLM recommendations (1--5 ordinal).}}
\label{tab:rec-rubric}
\scriptsize
\setlength{\tabcolsep}{4pt}
\renewcommand{\arraystretch}{1.15}
\revtablebg
\begin{tabularx}{\columnwidth}{|p{1.20cm}|X|p{0.80cm}|}
\hline
\textbf{Criterion} & \textbf{Anchors (1 / 3 / 5)} & \textbf{Scale} \\
\hline
Relevance &
1: Generic; weak linkage to diagnosed gaps. \newline
3: Mostly on-domain; partially tied to gaps. \newline
5: Precisely targets evidenced weaknesses (score/tier + salient signals). &
1--5 \\
\hline
Actionability &
1: Not actionable; lacks implementable steps. \newline
3: Some steps but missing feasibility/prioritization. \newline
5: Specific, feasible actions with clear levers and prioritization. &
1--5 \\
\hline
Faithfulness &
1: Adds unsupported facts or contradicts inputs. \newline
3: Minor extrapolation; mostly consistent with inputs. \newline
5: Strictly evidence-bounded; no unverifiable claims; consistent with inputs. &
1--5 \\
\hline
\end{tabularx}
\end{table}

\begin{table}[!t]
\centering
\caption{LLM-generated responses for selected countries in Group 2}
\label{tab:llm_suggestions}
\scriptsize
\setlength{\tabcolsep}{3pt} 
\renewcommand{\arraystretch}{1.12}

\begin{tabularx}{\columnwidth}{|p{0.65cm}|p{0.85cm}|p{1.15cm}|X|}
\hline
\textbf{Nation} & \textbf{Score} & \textbf{Feedback} & \textbf{Suggestion} \\
\hline
BG & 6.751 & Scored only 6.75/10.\newline Fair but can improve. &
Bulgaria should focus on enhancing incentives for renewable energy investment and improving energy efficiency in industrial sectors. The government needs to develop financial and technical support policies to promote green transition and strengthen international cooperation to learn and attract resources. \\
\hline
LV & 7.280 & Scored only 7.28/10.\newline Fair but can improve. &
Latvia should enhance policies supporting green businesses and promote investment in renewable energy. The government should consider providing financial and technical incentives to encourage the shift to more sustainable technologies and practices. \\
\hline
RS & 6.466 & Scored only 6.47/10.\newline Fair but can improve. &
Russia needs to improve the effectiveness of current policies and increase investment in green infrastructure. The government should promote renewable energy projects and implement energy-saving programs for households and businesses. \\
\hline
SK & 7.279 & Scored only 7.28/10.\newline Fair but can improve. &
Slovakia should strengthen financial support policies for renewable energy projects and improve energy efficiency in buildings. Also, SK should enhance international cooperation to learn from best practices and attract investment in advanced green technologies. \\
\hline
US & 6.996 & Scored only 7.00/10.\newline Fair but can improve. &
US is recommended to focus on increased investment in renewable energy projects and green infrastructure development. Stronger incentive policies are also needed to encourage businesses and individuals to adopt sustainable solutions, thereby enhancing the nation’s position in the global green market. \\
\hline
\end{tabularx}
\end{table}

\section{Conclusion} \label{sec-conclusion}
This study introduces an innovative AI-driven framework for assessing the ESG performance of European SMEs using data from the Flash Eurobarometer FL549 survey. The framework leverages AI agents to automates ESG benchmarking and evaluate sustainability metrics across four dimensions, with outputs showing minimal deviations from expert-validated baselines. This proposed approach has practical implications for policymakers by enabling targeted interventions aligned with the \textit{European Green Deal}. From a theoretical perspective, the study demonstrates the efficacy of human-AI collaboration in evidence-based policymaking, offering a scalable and resource-efficient solution for systematic ESG monitoring that addresses real-world capacity limitations in the SME sector.

\bibliographystyle{IEEEtran}
\bibliography{IEEEtran}

\end{document}